\pdfoutput=1
\documentclass{article}

\usepackage[main, final]{neurips_2026}

\usepackage[utf8]{inputenc}
\ifdefined\LaTeXML\else\usepackage[T1]{fontenc}\fi
\usepackage{graphicx}
\usepackage{float}
\usepackage{hyperref}
\usepackage{url}
\usepackage{booktabs}
\usepackage{amsmath}
\usepackage{amssymb}
\usepackage{amsfonts}
\usepackage{nicefrac}
\usepackage{microtype}
\usepackage{subcaption}
\usepackage{cleveref}
\usepackage{xcolor}
\hypersetup{pdftitle={BrickFlow: Connectivity-Guided Brick Reconstruction}, pdfauthor={Peter Kulits, Cordelia Schmid}}

\crefname{figure}{Fig.}{Figs.}
\crefname{table}{Tab.}{Tabs.}
\crefname{equation}{Eq.}{Eqs.}
\crefname{section}{Sec.}{Secs.}
\crefname{subsection}{Sec.}{Secs.}
\crefname{subsubsection}{Sec.}{Secs.}
\Crefname{figure}{Fig.}{Figs.}
\Crefname{table}{Tab.}{Tabs.}
\Crefname{equation}{Eq.}{Eqs.}
\Crefname{section}{Sec.}{Secs.}
\Crefname{subsection}{Sec.}{Secs.}
\Crefname{subsubsection}{Sec.}{Secs.}

\title{{BrickFlow}: Connectivity-Guided Brick Reconstruction}

\author{%
  Peter Kulits$^{1,2}$ \qquad Cordelia Schmid$^{1}$ \\
  $^{1}$Inria, \'Ecole Normale Sup\'erieure, CNRS, PSL Research University \\
  $^{2}$Max Planck Institute for Intelligent Systems, T\"ubingen \\
}

\crefname{appendix}{App.}{Apps.}
\Crefname{appendix}{App.}{Apps.}
\begin{document}
\raggedbottom

\maketitle

\begin{figure}[H]
  \centering
  \vspace{-0.5cm}
  \ifdefined\LaTeXML
    \includegraphics[width=\linewidth]{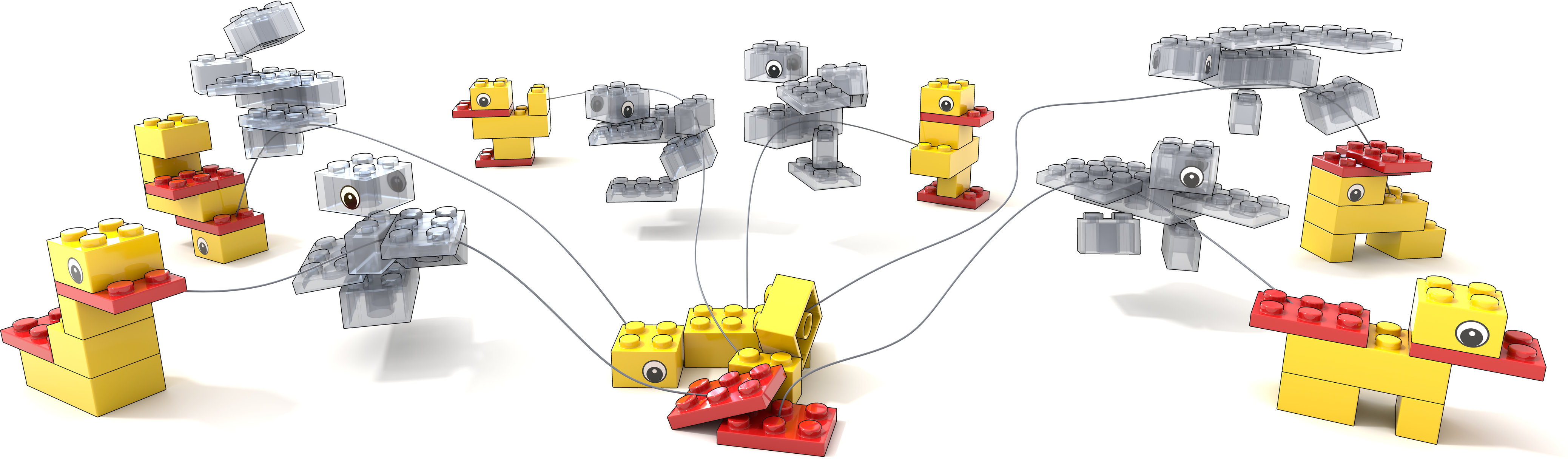}
    \includegraphics[width=\linewidth]{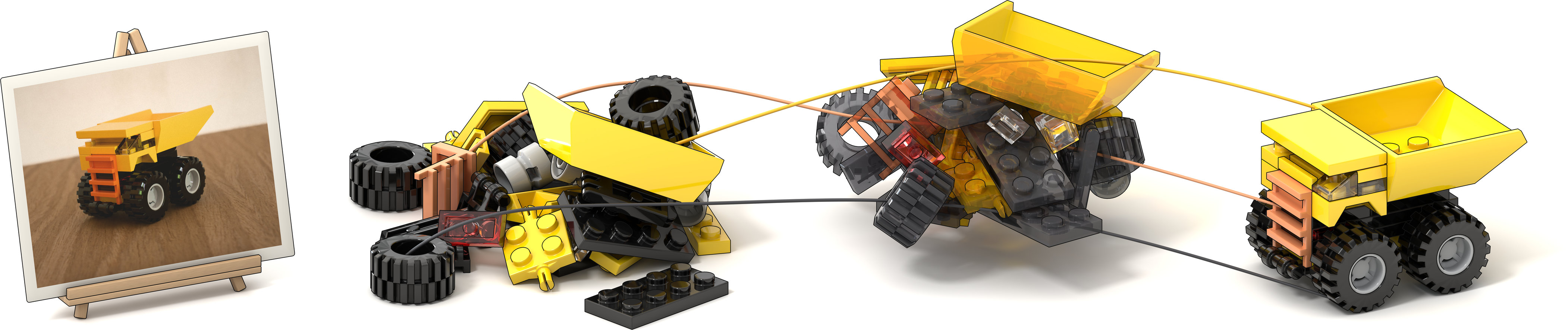}
  \else
    \makebox[\linewidth][l]{\includegraphics[width=1.013916\linewidth]{fig/images/teaser_duck.jpg}}\\[4pt]
    \makebox[\linewidth][l]{\includegraphics[width=1.026611\linewidth]{fig/images/teaser_truck.jpg}}
  \fi
  \caption{\textbf{Top:} We present BrickFlow, an SE(3) flow-matching model trained to take in a known set of LEGO pieces (pile) with noisy poses (translucent) and denoise them into some valid structure.
    \ifdefined\LaTeXML\newline\fi\textbf{Bottom:} Using a procedural dataset 1\% the size, we adapt it for image-conditioned reconstruction.
  }
  \label{fig:teaser}
\end{figure}

\begin{abstract}
  Given a set of LEGO parts and an image of an assembled object, our task is to infer the pose of each part to assemble the object.
  In contrast with traditional 3D reconstruction, doing so requires satisfying discrete physical constraints, which provide structure but render the problem combinatorial and highly sensitive to small pose errors.
  To learn a prior over how parts fit together, we train a set-conditioned $SE(3)$ flow model on procedurally generated assemblies.
  We then finetune this model to inject image conditioning, using 1\% as much data.
  However, while the learned flow captures global structure, it does not enforce connectivity.
  To address this, we introduce an analytic connectivity-field guidance to pull predictions toward connected assemblies at test time.
  Through this, we demonstrate improved build quality with model size, pretraining, and guidance.
  We make our models and data available for research purposes at
  \url{https://kulits.github.io/BrickFlow/}
\end{abstract}

\section{Introduction}\label{sec:intro}
Imagine opening a new set of LEGO to find a box full of pieces, but no instructions.
All you have is the picture on the front of the box.
You can identify a number of parts visible on the surface, but the image only reveals the outside.
The internal structure must be inferred: which bricks support which, what fills the spaces you cannot see.
Every piece must end up with a precise position and orientation satisfying discrete connector-defined constraints.
This is the problem we address: given an inventory of parts and a single image of the completed assembly, recover the 6D pose of each piece (see \cref{fig:teaser}).

While sharing some of the ambiguities of traditional monocular 3D reconstruction -- such as from occlusion -- the task presents unique challenges.
Though it is possible to computationally verify the validity of a given assembly, figuring out how to go from parts to structure requires understanding not just how pieces fit together, but how they fit together jointly, over possibly many at once.
Contributing to this complexity, our assembly domain spans thousands of parts which connect in diverse ways.

We address this by learning a partset-conditioned generative assembly prior.
To obtain sufficient supervision, we generate large-scale procedural structure data using a typed connector system~\citep{kulits2026bricknet} defined over LDraw~\citep{ldraw} parts, starting from randomly sampled pieces and iteratively placing new ones through compatible connectors.
We use these to train an $SE(3)$ flow model that takes an unordered inventory together with noisy poses and removes noise.

The purpose of this stage is not to understand object semantics, but to learn the geometry of assembly: which parts can connect, how they tend to align, and how local connections compose into global structure.
Rather than being human-designed or semantically meaningful, these assemblies provide constraint-rich samples from the physical grammar of LEGO assembly.
Pretraining on this data amortizes search over connector-valid configurations into a reusable prior on assembly and structure.

We then adapt the assembly prior to perform partset-conditioned visual reconstruction.
For this, we render a smaller set of procedural assemblies to create structure--image pairs.
From the images, we extract features from a frozen visual encoder and finetune the flow model with injected cross-attention.

Although pretraining captures useful global assembly structure, learned predictions alone do not guarantee precise connectivity.
Small translational and rotational errors can leave intended connections slightly misaligned.
To address this, we introduce connectivity-field guidance, an analytic test-time energy defined over poses and connector geometry.
The field is zero on exactly connected assemblies, and its gradient pulls predictions toward connected structure.
Following diffusion posterior sampling (DPS)~\citep{chung2023dps}, we condition on connectedness with the field as the negative log-likelihood, correcting poses in late steps along its gradient at the model's predicted final assembly.

We evaluate our model on both inventory-conditioned assembly and image-conditioned reconstruction.
Beyond pose accuracy, we evaluate assembly and connection validity.
We demonstrate that assembly pretraining improves structural understanding, that task performance scales with model capacity, and that connectivity-field guidance improves the validity of assemblies when applied at inference time.

Our core contributions are as follows:
\begin{enumerate}
  \item We train a set-conditioned $SE(3)$ flow model on procedurally generated LEGO assemblies to learn an adaptable assembly prior
  \item We adapt this prior to single-image reconstruction using a fraction the amount of data
  \item We introduce connectivity-field guidance, an analytic energy that injects known connector structure at test time, pulling predictions toward connected assemblies without retraining
\end{enumerate}

\section{Related Work}
\label{sec:related}
\noindent\textbf{Brick Assembly.}
Prior work on brick assembly has addressed several specific subproblems.
Legolization methods convert target geometry into brick layouts on discrete grids~\citep{luo2015legolization,lennon2021image2lego}.
BrickGPT~\citep{pun2025legogpt} finetunes a language model to autoregressively place bricks, but predicts voxel positions over a small block vocabulary; LEGO-Maker~\citep{ge2025legomaker} creates LEGO models from an image over a similarly restricted part set.
For reconstruction, \citet{tattersall2020reconstructing} recover LEGO models from 24 calibrated views via space carving and cuboid fitting, but handle only four rectangular brick types; \citet{wang2022mepnet} infer per-step brick poses from instruction manuals but address sequential single-part placement rather than joint assembly.
BrickNet~\citep{kulits2026bricknet} introduce a typed connector system and graph-based generation over the full LDraw part library and finetune a language model to autoregressively produce brick structures from text.
Each step places a part through a connection to the existing structure, so its outputs are connected by construction, and it takes no image or inventory.
BrickBench~\citep{kulits2026brickbench} evaluates coding agents on text-conditioned design of buildable assemblies with up to 4000 parts using the same library and tests them under different constraints.
Apart from the per-step placement of \citet{wang2022mepnet}, these methods choose which parts to use, and none is provided a fixed set of parts.
Providing an inventory constrains the structure without determining it, as the same parts can build many assemblies.
As it must be used exactly, a model cannot omit a part that does not fit and must connect every part.
No prior work addresses inventory-conditioned assembly or reconstruction over the full part library.

\noindent\textbf{3D Part Assembly.}
Generative part assembly methods predict rigid poses for a set of parts to form a complete object~\citep{huang2020generative,li2020singleimage}, with recent work using score-based~\citep{cheng2023scorepa}, diffusion~\citep{scarpellini2024diffassemble,zhao2025assembler}, hierarchical~\citep{du2024partwhole}, and flow-based~\citep{li2025garf,sun2025rectifiedpointflow} formulations.
GARF~\citep{li2025garf} and Rectified Point Flow~\citep{sun2025rectifiedpointflow} are 3D-only, though Assembler~\citep{zhao2025assembler} is additionally image-conditioned.
While these are architecturally similar (we employ a standard DiT~\citep{Peebles_2023_ICCV}), they differ in task, and are designed for arbitrary geometry encoded as sampled points.

\noindent\textbf{SE(3) Flow Matching and Guided Sampling.}
The flow matching paradigm~\citep{lipman2023flowmatching,liu2022rectifiedflow} has been extended to $SE(3)$ for protein structure generation~\citep{bose2024foldflow,yim2023frameflow,campbell2024generative}, where, as in our setting, structures are built from a fixed vocabulary of units rather than arbitrary parts.
Diffusion posterior sampling (DPS)~\citep{chung2023dps} corrects each denoising step using gradients of a measurement likelihood; classifier~\citep{dhariwal2021diffusion} and classifier-free~\citep{ho2022classifierfree} guidance follow the same principle with learned signals.
Our connectivity-field guidance applies the DPS correction with an analytic energy over known connector geometry as the negative log-likelihood.
No prior work integrates connectivity differentiably as test-time guidance in an assembly task.

\section{Methods}\label{sec:methods}

\subsection{Modeling}\label{ssec:modeling}
We represent LEGO assemblies using the LDraw~\citep{ldraw} modeling system, which consists of digital replicas of thousands of parts.
In LDraw, a model is defined by a set of piece instances, each having type, material, and $SE(3)$ pose.
It does not directly model part connectivity, for which we employ the BrickNet~\citep{kulits2026bricknet} typed connector system, an annotation layer of local connector frames over LDraw parts which defines how pieces connect to one another.
These connection interfaces include stud--hole, hinge, axle, ball joint, and fixed connectors, each defining a family of allowed relative transforms between two parts.
Stud-type connectors, for instance, allow for one degree of rotational freedom about an interface.
Given an LDraw assembly, BrickNet parses posed parts into a graph where structure is defined by pieces as nodes and edges of connectivity.
In this work, we use LDraw as the source of part geometry, while the BrickNet connector system is used to evaluate the validity of an assembly and integral to the production of procedural training data.

\subsection{ProcBrick}\label{ssec:data}
\begin{figure}[t]
  \centering
  \includegraphics[width=\linewidth]{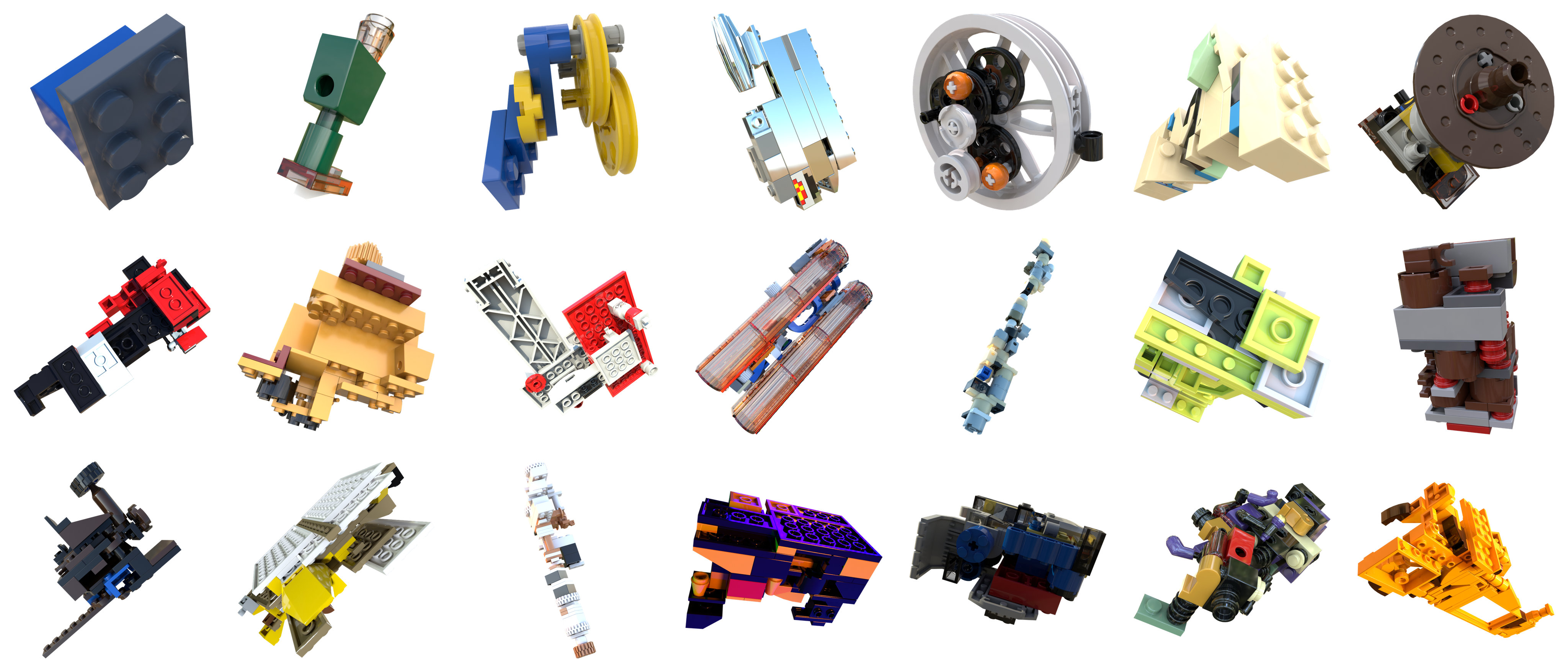}
  \caption{
    \textbf{ProcBrick samples.}
    We provide sample renders from ProcBrick of increasing complexity.
    The observable differences in shading are due to random HDRI selection done at the time of rendering.
  }
  \label{fig:data-sample-renders}
  \vspace{-0.15cm}
\end{figure}

To train our assembly prior, we introduce ProcBrick, a procedural generator of LEGO assemblies, and use it to synthesize a dataset of 100 million structures of 100 parts each.
Parts are drawn from 6,339 geometry classes, which we use to group LDraw parts with identical colliders and connector interfaces.
Starting with a random seed piece, we iteratively append pieces onto the structure.
Rather than sample naively in space, we attempt placements at random compatible connector sites and reject those that collide.
However, naive rejection sampling favors sparse assemblies: placements that extend into open space rarely collide, while those among existing parts, where further connections would form, do so more frequently.
We instead stochastically accept collision-free placements to favor dense packing and, for each new part, numbers of connector joins and connected neighbors within the range of human-designed models, so that assemblies are compact and multiply connected.
As sampling is autoregressive, each sequence prefix defines a valid structure and lengths can be varied in pretraining.
For image-conditioned finetuning, we generate an additional one million ProcBrick structures, 1\% as many as for pretraining, and render them with Blender.
Rather than consistently include 100 parts, we balance counts between 2 and 100.
For rendering, each part is assigned a printed or plain variant of its geometry class and a material, giving 12,663 distinct parts.
We fit part and material frequencies to human-designed models (see \cref{sec:supp_data}).
See \cref{fig:data-sample-renders} for render samples.
Although the procedural assemblies are not individually realistic as sets a human might construct, our intent is not to model semantics but to span the space of valid structure to achieve broad support.

\subsection{Problem}\label{ssec:problem}
We formulate our task as 6-DoF pose prediction over an unordered multiset of known rigid parts:
\begin{equation}
  \mathcal{P} = \{(p_i, m_i)\}_{i=1}^N,
\end{equation}
where $p_i$ denotes the part type and $m_i$ is the material.
The goal is to predict one transform per part,
\begin{equation}
  X = \{T_i\}_{i=1}^N,
  \qquad
  T_i \in SE(3),
\end{equation}
such that the transformed parts form the object depicted in a provided image $I$.
Poses are predicted with the assembly centered at its mass-weighted centroid but in an arbitrary global orientation, so we evaluate up to a global rigid transform.
As the inventory may contain duplicate parts, we do not fix ordering.
Parts may have rotational symmetries.
Training and evaluation therefore account for these three equivalences: global rigid transform, duplicate-part permutation, and intrinsic part symmetry.

We consider two settings.
In inventory-conditioned assembly, which we refer also to as pretraining, the model receives only the part inventory and is tasked with predicting some valid assembly.
In image-conditioned reconstruction, the model is provided an image and predicts the depicted assembly.

\noindent\textbf{Pose distance.}
We measure distance between two poses of a part by how far material moves.
With pose $T=(R,\mathbf{x})$ placing the part's center of mass at $\mathbf{x}$, the mean squared displacement of material is
\begin{equation}
  d^2(T, T') = \|\mathbf{x} - \mathbf{x}'\|^2 + \operatorname{tr}\!\big((R - R')\,\Sigma\,(R - R')^\top\big),
  \label{eq:metric}
\end{equation}
where $\Sigma$ is the part's second moment of mass per unit mass~\citep{bregier2018pose}.
Translation and rotation are thus measured in the same unit, LDraw units (LDU; 20\,LDU is the width of a stud).

\subsection{SE(3) Flow Matching}\label{ssec:flow_matching}
We train BrickFlow as a flow model over sets of rigid poses.
Following flow-matching conventions, $t{=}0$ denotes the noise endpoint and $t{=}1$ the data endpoint.
Let $X_1$ denote a target assembly and $X_0$ an initial noisy pose set.
Noise translations are drawn from an isotropic Gaussian, shifted to zero mass-weighted centroid, with a standard deviation that depends only on the part count $N$ and is fitted to the spread of $N$-part training assemblies; noise rotations are drawn uniformly from $SO(3)$.
Because the input inventory is unordered and parts may be symmetric, we pair $X_0$ with $X_1$ by approximately minimizing the total mass movement $\sum_i \mu_i\, d^2(T_i^0, T_{\pi(i)}^1)$, with part masses $\mu_i$, over permutations $\pi$ of duplicate parts, part symmetries, and a global rotation~\citep{klein2023equivariant}.
For each training example, we sample $t$ from a logit-normal distribution~\citep{esser2024scaling} and construct an interpolated pose set $X_t$ between $X_0$ and $X_1$.
Translations are then interpolated linearly,
\begin{equation}
  \mathbf{x}_i(t) = (1-t)\mathbf{x}_i^0 + t\mathbf{x}_i^1,
\end{equation}
and rotations are interpolated along the geodesic on $SO(3)$,
\begin{equation}
  R_i(t) = R_i^0 \exp\left(t \log\left((R_i^0)^\top R_i^1\right)\right).
\end{equation}
These paths have constant velocities $\mathbf{v}_i = \mathbf{x}_i^1 - \mathbf{x}_i^0$ and $\boldsymbol{\omega}_i = \log\big((R_i^0)^\top R_i^1\big)^\vee$, the rotation vector of the relative rotation.
The model predicts the translational and angular velocity of each part, $(\hat{\mathbf{v}}_i, \hat{\boldsymbol{\omega}}_i) = f_\theta(X_t, t, \mathcal{P}, I)$, where $I$ and the materials are omitted during structure-only pretraining.
We train it with the infinitesimal form of \cref{eq:metric},
\begin{equation}
  \mathcal{L} = \|\hat{\mathbf{v}}_i - \mathbf{v}_i\|^2 + w(t)\,(\hat{\boldsymbol{\omega}}_i - \boldsymbol{\omega}_i)^\top \tfrac{J_i}{\mu_i}\,(\hat{\boldsymbol{\omega}}_i - \boldsymbol{\omega}_i),
\end{equation}
averaged over parts, where $J_i$ is the part's inertia tensor and $w(t) = \big((1-t)/(1-\min(t, 0.9))\big)^2$ down-weights rotation errors near the data, as in FrameFlow~\citep{yim2023frameflow}.
The velocities imply a clean estimate $\hat{X}_1$, with $\hat{\mathbf{x}}_i^1 = \mathbf{x}_i(t) + (1-t)\hat{\mathbf{v}}_i$ and $\hat{R}_i^1 = R_i(t)\exp\big((1-t)\hat{\boldsymbol{\omega}}_i^\wedge\big)$.

At inference, we integrate translations with 50 Heun steps~\citep{karras2022elucidating}.
Following FrameFlow and MultiFlow~\citep{yim2023frameflow,campbell2024generative}, rotations instead advance a fraction $\kappa\,\Delta t$ of the geodesic toward $\hat{R}_i^1$ with a fixed rate $\kappa{=}25$, rather than the $\Delta t/(1{-}t)$ implied by the training interpolant, which roughly halves unguided validity (\cref{sec:supp_guidance_sweep}).
While the loss does not enforce that predicted parts connect, the model learns structure from seeing training samples that are connected.

\subsection{Architecture}\label{ssec:architecture}
\begin{figure*}[t]
  \centering
  \ifdefined\LaTeXML
    \begin{subfigure}[b]{0.49\linewidth}
      \includegraphics[width=\linewidth]{fig/images/pipeline_pretrain.svg}
      \caption{Structure-only pretraining}
      \label{fig:pipeline-pretrain}
    \end{subfigure}\hfill
    \begin{subfigure}[b]{0.49\linewidth}
      \includegraphics[width=\linewidth]{fig/images/pipeline_finetune.svg}
      \caption{Image-conditioned finetuning}
      \label{fig:pipeline-finetune}
    \end{subfigure}
  \else
    \begingroup
    \newlength{\pipepanelwidth}
    \newlength{\pipepanelheight}
    \setlength{\pipepanelwidth}{\dimexpr0.5\linewidth-5pt\relax}
    \settoheight{\pipepanelheight}{\includegraphics[width=\pipepanelwidth,trim=16.575 15.011 10.555 6.099]{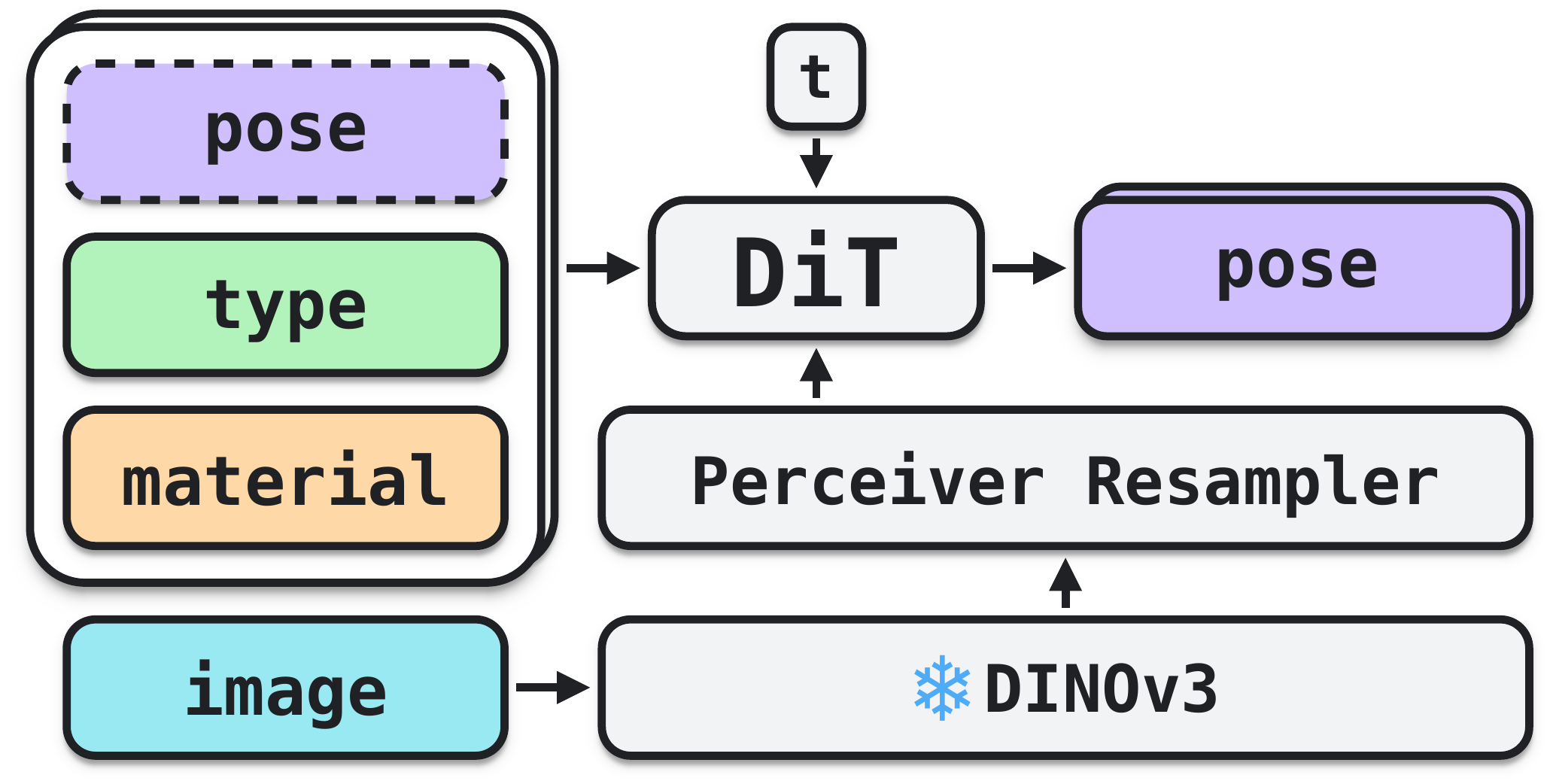}}
    \begin{subfigure}[b]{\pipepanelwidth}
      \parbox[t][\pipepanelheight][t]{\linewidth}{\vspace*{0.114794\linewidth}%
        \includegraphics[width=\linewidth,trim=16.575 23.327 10.555 6.099]{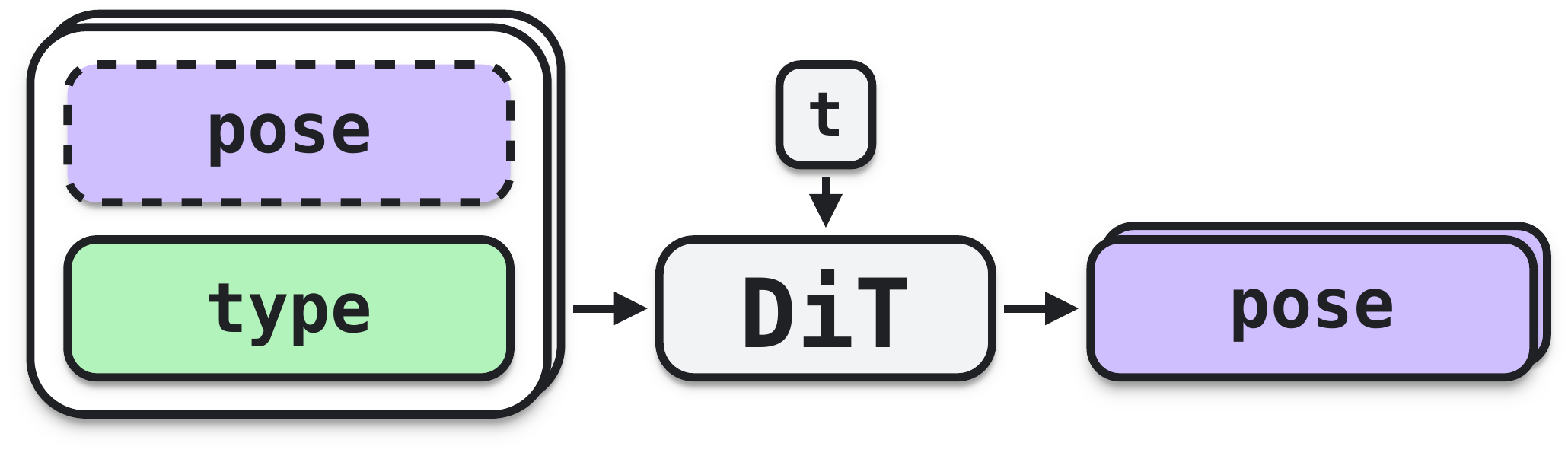}}
      \caption{Structure-only pretraining}
      \label{fig:pipeline-pretrain}
    \end{subfigure}\hfill
    \begin{subfigure}[b]{\pipepanelwidth}
      \includegraphics[width=\linewidth,trim=16.575 15.011 10.555 6.099]{fig/images/pipeline_finetune.pdf}
      \caption{Image-conditioned finetuning}
      \label{fig:pipeline-finetune}
    \end{subfigure}

    \endgroup
  \fi
  \caption{\textbf{Pipeline.}
    (a) BrickFlow denoises poses given known part types, conditioned on \hbox{flow time $t$.}
    (b) Finetuning adds materials and conditioning on frozen DINOv3 features with a Perceiver resampler.}
  \label{fig:pipeline}
  \vspace{-0.01cm}
\end{figure*}

We use a standard diffusion-transformer (DiT)~\citep{Peebles_2023_ICCV} backbone without $SE(3)$-equivariant layers (\cref{fig:pipeline}).
Prior work on $SE(3)$ flow matching for proteins~\citep{yim2024motif} uses invariant point attention (IPA) layers from AlphaFold~2~\citep{jumper2021alphafold}, but with data augmentation, non-equivariant models can match the data efficiency of equivariant ones given sufficient training~\citep{brehmer2025does}; AlphaFold~3~\citep{abramson2024alphafold3} achieves strong results with a non-equivariant architecture and data augmentation, which we follow.
Each input token represents a part: the token embedding is the sum of a learned part-type embedding, linear encodings of the current pose, and optionally a learned material embedding.
As the input is unordered, no positional encoding is used.
In the reconstruction experiment, we inject image conditioning through decoupled cross-attention~\citep{ye2023ipadapter} in each of the DiT blocks from part tokens to frozen DINOv3-L~\citep{dinov3} features (CLS and patch tokens) from layer 18 of 24, compressed by a Perceiver resampler~\citep{jaegle2021perceiver,alayrac2022flamingo}.
A keypoint head additionally predicts where each part's center of mass projects into the image, trained with a $\beta$-NLL loss~\citep{seitzer2022pitfalls}.
We drop the image for 10\% of training examples and sample with classifier-free guidance~\citep{ho2022classifierfree} of strength 2, handing off to the pretrained model for the final steps ($t \geq 0.85$).

\subsection{Connectivity Field}\label{ssec:connectivity_field}
While large-scale pretraining teaches the model a useful global assembly prior, the learned flow does not explicitly enforce connector alignment.
Small positional errors or drifts can leave two parts misaligned, leading to an assembly where the predicted structure might be qualitatively close, but disconnected.
This is a natural failure mode for direct pose prediction in $SE(3)^N$: the set of connector-consistent configurations has measure zero in the full space.
To address this, we introduce a connectivity-field guidance, a scalar energy defined over joint pose whose gradient nudges estimates toward connected configurations.
We apply it at test time during sampling.
The field encodes the same constraints that define a valid connection, but as a soft energy: enforcing them as hard constraints would presuppose the connection graph, which is precisely what is inferred jointly with the poses.

\begin{figure}[t]
  \centering
  \ifdefined\LaTeXML
    \begin{subfigure}[b]{0.1992\linewidth}
      \includegraphics[width=\linewidth]{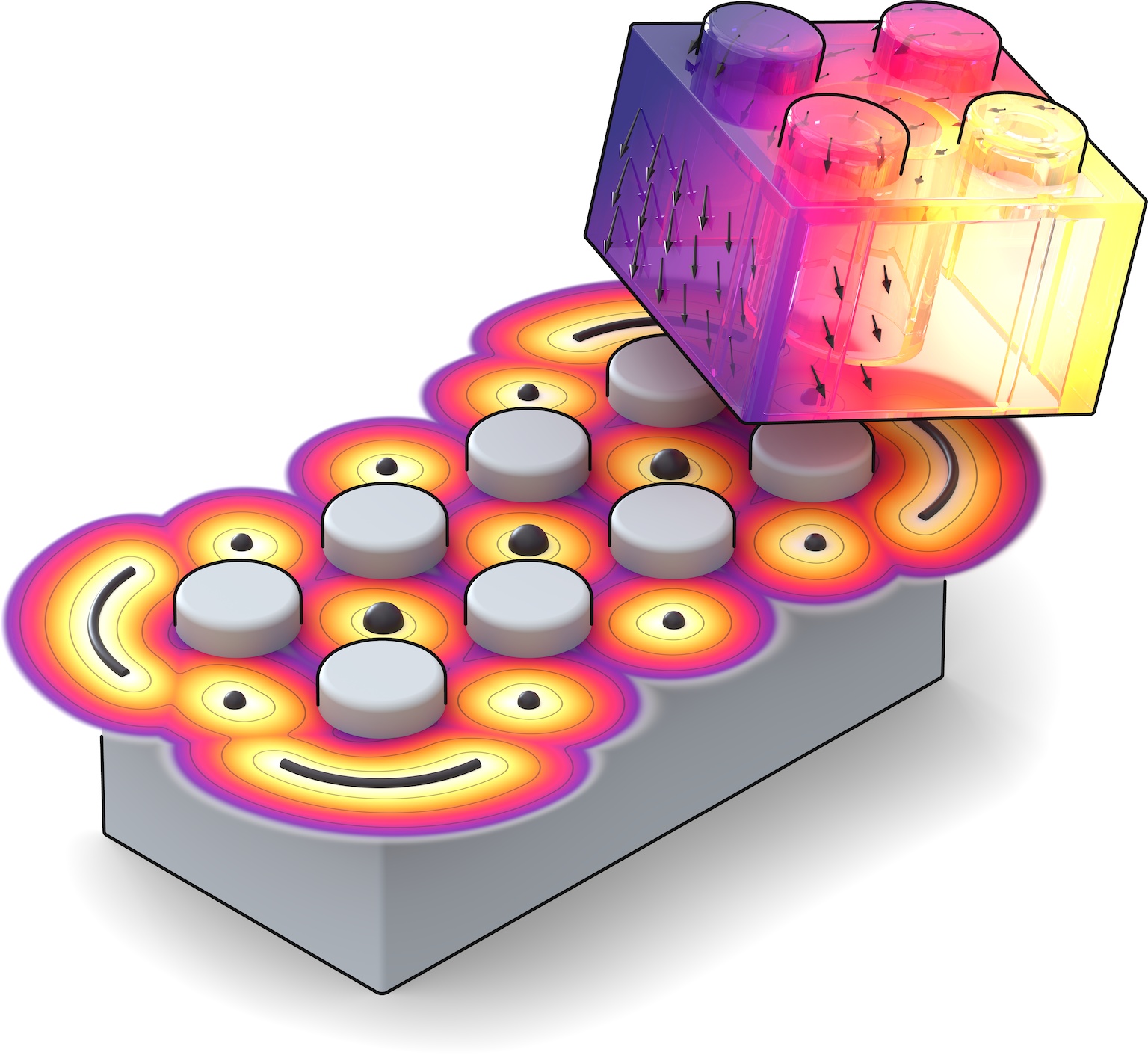}
      \caption{Pair energy}
      \label{fig:connectivity-field-pair}
    \end{subfigure}\hfill
    \begin{subfigure}[b]{0.337\linewidth}
      \includegraphics[width=\linewidth]{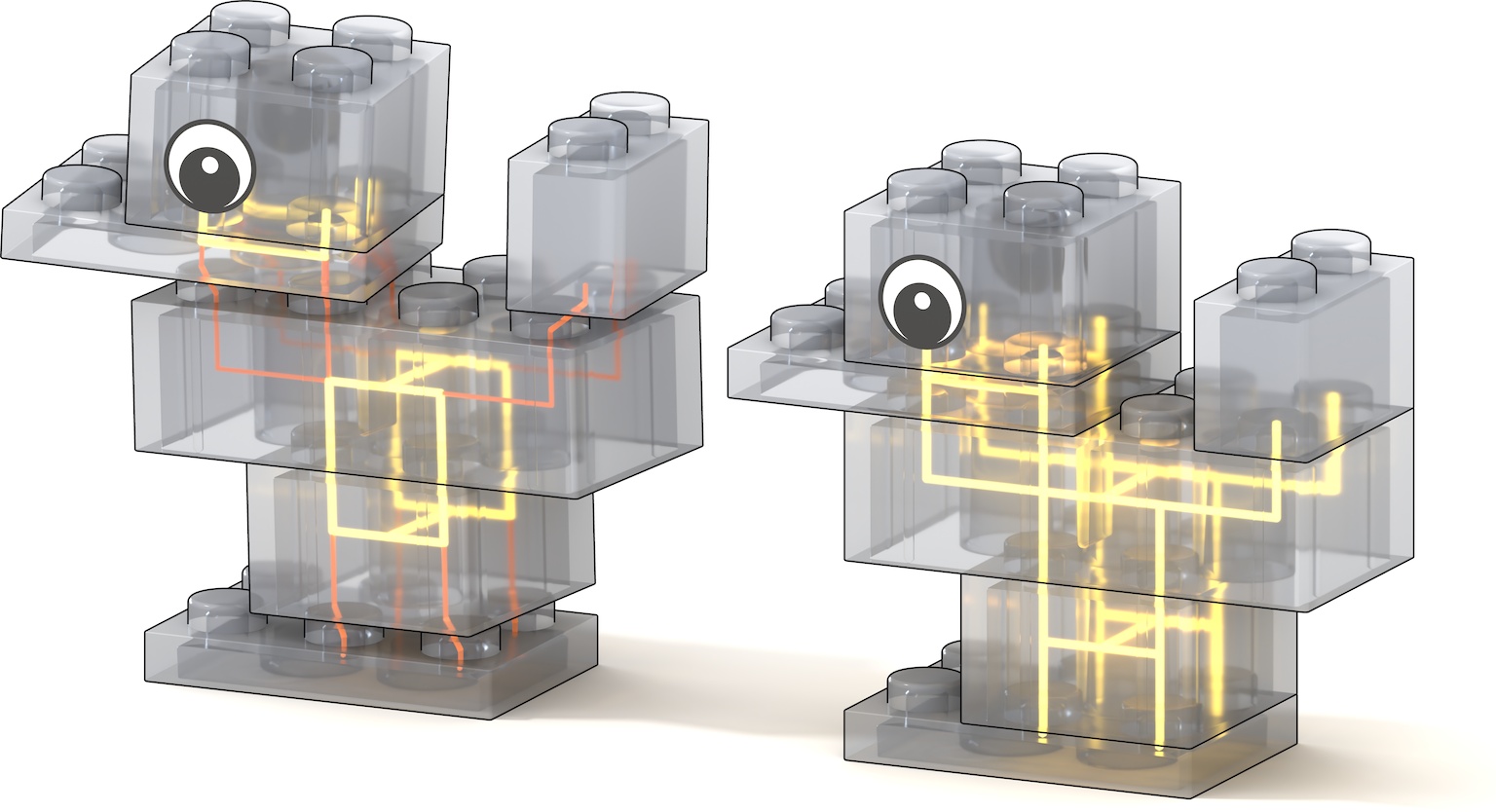}
      \caption{Network}
      \label{fig:connectivity-field-network}
    \end{subfigure}\hfill
    \begin{subfigure}[b]{0.4398\linewidth}
      \includegraphics[width=\linewidth]{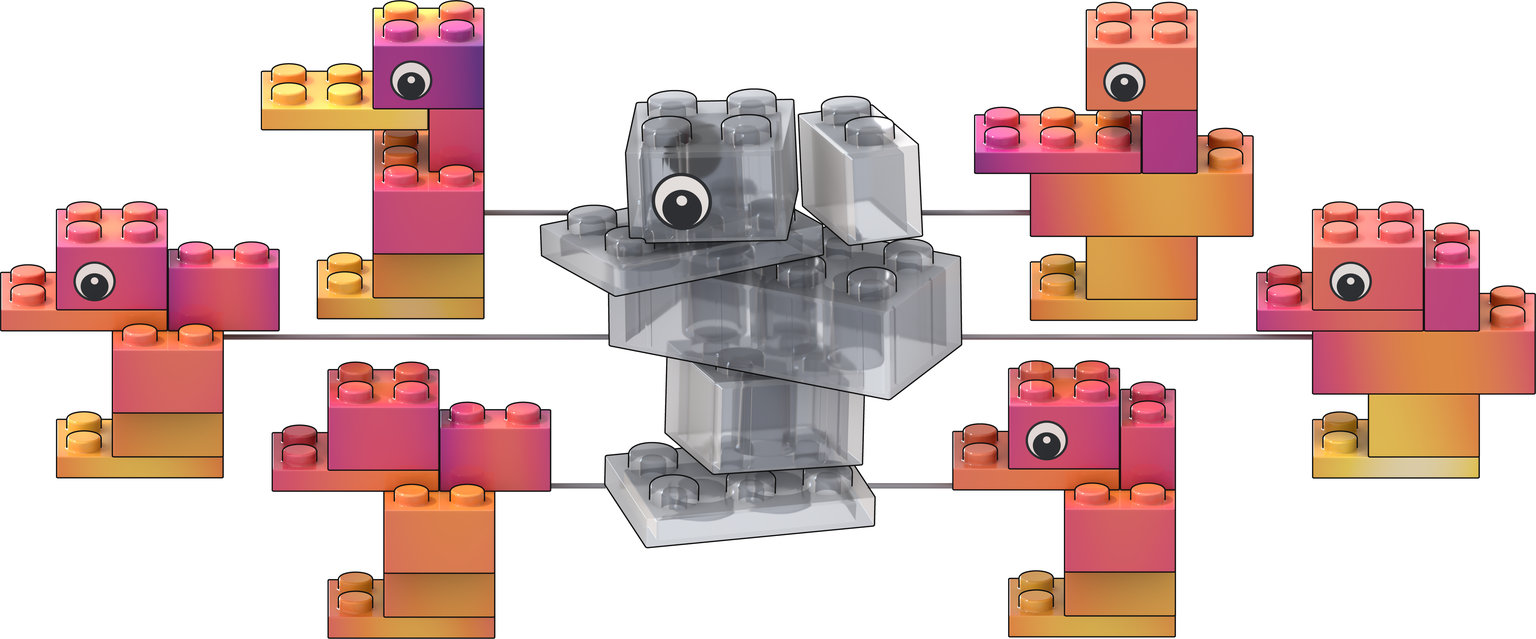}
      \caption{Energy minima}
      \label{fig:connectivity-field-completions}
    \end{subfigure}
  \else
    \newlength{\cfgap}\setlength{\cfgap}{0.012\linewidth}
    \newlength{\cfh}\setlength{\cfh}{\dimexpr(\linewidth-2\cfgap)*1000/5334\relax}
    \newlength{\cfw}
    \settowidth{\cfw}{\includegraphics[height=\cfh]{fig/images/field_pair.jpg}}%
    \begin{subfigure}[b]{\cfw}
      \includegraphics[height=\cfh]{fig/images/field_pair.jpg}
      \caption{Pair energy}
      \label{fig:connectivity-field-pair}
    \end{subfigure}\hspace{\cfgap}%
    \settowidth{\cfw}{\includegraphics[height=\cfh]{fig/images/field_circuit.jpg}}%
    \begin{subfigure}[b]{\cfw}
      \includegraphics[height=\cfh]{fig/images/field_circuit.jpg}
      \caption{Network}
      \label{fig:connectivity-field-network}
    \end{subfigure}\hspace{\cfgap}%
    \settowidth{\cfw}{\includegraphics[height=\cfh]{fig/images/field_ring.jpg}}%
    \begin{subfigure}[b]{\cfw}
      \includegraphics[height=\cfh]{fig/images/field_ring.jpg}
      \caption{Energy minima}
      \label{fig:connectivity-field-completions}
    \end{subfigure}
  \fi
  \caption{
    \textbf{Connectivity field.}
    (a) The energy $E$ of a two-part assembly as a function of where the upper brick is placed on the lower one.
    It is zero (ink) exactly at valid seats ($c_h = 0$): isolated points where the brick is held rigidly, and arcs where it connects to a single stud and can swivel.
    The upper brick is colored by how far its material must move to reach a valid seat (\cref{eq:metric}), with needles showing the direction.
    (b) $E$ models each assembly as an electrical network: parts are nodes, and each candidate seat $h$ adds a conductance $g_h$ between two parts (drawn as wires, brighter for higher $g_h$).
    Seated connections join every part of the valid duck (right); the disordered estimate (left) consists of seated groups joined only through weak, unseated candidates, so its effective resistances $\Omega_{ab}$, and hence $E$, are high.
    (c) A noisy estimate (center) and six nearby valid structures built from the same parts, colored by how far their material moves from the estimate.
    $E = 0$ on each of them.
  }
  \label{fig:connectivity-field}
  \vspace{-0.03cm}
\end{figure}

\noindent\textbf{Seats.}
For each candidate seat $h$, a pair of compatible connectors on parts $a$ and $b$, we measure how far the parts are from a valid seat by $c_h \geq 0$, the mean squared material displacement (\cref{eq:metric}) from their relative pose to the nearest allowed relative transform of the connector family, minimized over its free coordinates (e.g., stud rotation, axle slide).
A perfect mate has $c_h = 0$.

\noindent\textbf{Network.}
We treat the assembly as an electrical network with parts as nodes (\cref{fig:connectivity-field}b).
Each candidate seat $h$ contributes conductance $g_h = w_h / (\rho + q(c_h))$, where $\rho$ is a small seat resistance, $q$ is zero for $c_h$ below a small tolerance and grows smoothly beyond it, and $w_h$ tapers as connectors move apart (12--16\,LDU).
The candidate seats between two parts combine as $G_{ab} = (\sum_h g_h^4)^{1/4}$, so the best one dominates, and a weak conductance between every pair of parts keeps the network connected.
Let $\Omega_{ab}$ denote the effective resistance~\citep{doyle1984random} between parts $a$ and $b$.
The energy is
\begin{equation}
  E(X) = \frac{1}{N-1} \sum_{a<b} \phi\big(\Omega_{ab} - (N-1)\rho\big),
\end{equation}
where $\phi$ is a smooth hinge that is zero for nonpositive arguments and positive otherwise (\cref{sec:supp_connectivity}).

\noindent\textbf{Properties.}
If exactly seated connections join all parts, every pair is linked by a path of at most $N{-}1$ resistances $\rho$, so $\Omega_{ab} \leq (N-1)\rho$ and both $E$ and its gradient vanish.
Additional conductances can only lower $\Omega_{ab}$ (Rayleigh's monotonicity law~\citep{doyle1984random}), so loose contacts and redundant links do not disturb a connected assembly.
In contrast, a detached group of parts has a high effective resistance to the rest and is pulled toward it through candidate seats in reach.

\subsection{Test-Time Guidance}\label{ssec:guidance}
We seek samples from the energy-tilted distribution
\begin{equation}
  \tilde{p}(X_1) \;\propto\; p(X_1)\,\exp\!\bigl(-\beta E(X_1)\bigr),
\end{equation}
which reweights the model's output distribution toward connected assemblies.
As $E \geq 0$, $\exp(-\beta E) \in (0, 1]$ for $\beta > 0$ is a valid likelihood of observing that the assembly is connected, so $\tilde{p}$ is the posterior given that observation.
Following DPS~\citep{chung2023dps}, we approximate sampling from $\tilde{p}$ by evaluating the energy at the clean estimate $\hat{X}_1(X_t)$ and, for $t \geq t_{\mathrm{start}}$, correcting the state:
\begin{equation}
  X_{t+\Delta t} \;\leftarrow\; X_{t+\Delta t} \;-\; \gamma\, M^{-1} \nabla_{X_t} E\bigl(\hat{X}_1(X_t)\bigr),
\end{equation}
where $\gamma$ absorbs $\beta$, the correction is applied as a twist about each part's center of mass, and $M$ weights each part's rotation by its squared radius of gyration.
We clip each part's correction to 1\,LDU and remove the mass-weighted mean translation, so the centroid stays fixed as in training.
We use $\gamma = 2$ for all models and $t_{\mathrm{start}} = 0.7$, or 0.85 for image-conditioned models, where the pretrained model takes the final steps; we select these on a subset of the ProcBrick evaluation set (\cref{sec:supp_guidance_sweep}).
On that subset, varying $\gamma$ from 1 to 4, the clip from 0.5 to 2\,LDU or removing it, or $t_{\mathrm{start}}$ from 0 to 0.8 changes validity by at most 2.1 points (\cref{tab:dps_choices}).
Each guided step adds one backward pass (\cref{sec:supp_timing}); no retraining or auxiliary networks are needed to apply the guidance at sampling time during inference.

\section{Experiments}
\label{sec:experiments}

\subsection{Setup}

\noindent\textbf{Models.}
We pretrain inventory-conditioned DiT-S/B/L (PT) on 100M procedural assemblies.
For image-conditioned (IC) reconstruction, we finetune each on 1M rendered assemblies (\cref{tab:training}).
For ablations, we additionally train IC DiT-S from random initialization and ablate the amount of data.

\noindent\textbf{Evaluation sets.}
We evaluate on 9,900 held-out ProcBrick assemblies, 100 for each part count from 2 to 100, and on the 512 models of the BrickNet~\citep{kulits2026bricknet} validation set, with 512$\times$512 input images.
We draw 100 samples.
\textit{DPS} uses connectivity-field guidance (\cref{ssec:guidance}).

\noindent\textbf{Metrics.}
Pose error alone does not show whether an assembly holds together, so we report connectivity and collisions alongside it.
\emph{Valid}: BrickNet's connectivity and collision checks~\citep{kulits2026bricknet}, applied to the whole assembly rather than per build step: the parsed connector graph has one component and no two parts collide by more than 1.1\,LDU ($\uparrow$); BrickNet's default of 1.0\,LDU gives nearly identical results (\cref{tab:full}).
\emph{Comp.~}and \emph{Max comp.}: number of connected components ($\downarrow$) and size of the largest one ($\uparrow$).
\emph{MM} (mass movement): root-mean-square material displacement (\cref{eq:metric}) to the ground truth in LDU, with every part weighted equally, after optimal global alignment and symmetry-aware assignment of interchangeable parts ($\downarrow$).
\emph{Coll.}: colliding part pairs per sample ($\downarrow$).
We report 95\% confidence intervals from a bootstrap over structures.

\begin{table}[t]
  \caption{\textbf{Evaluation.}
    Procedural (ProcBrick) and human-designed (BrickNet) assemblies, 100 samples per structure.
    PT DiT-L receives only the inventory.
    Valid: 95\% CI is shown in brackets.}
  \label{tab:main}
  \centering
  \small
  \setlength{\tabcolsep}{4pt}
  \begin{tabular}{@{}l l r r r r r@{}}
    \toprule
    Model & Guid. & Valid (\%) $\uparrow$ & Comp.\ $\downarrow$ & Max comp.\ $\uparrow$ & MM $\downarrow$ & Coll.\ $\downarrow$ \\
    \midrule
    \multicolumn{7}{l}{\textit{ProcBrick}} \\
    PT DiT-L & -- & 10.3 {\scriptsize[9.9, 10.8]} & 7.0 & 33.5 & 72.6 & 18.6 \\
    & DPS & 13.5 {\scriptsize[12.9, 14.0]} & 2.9 & 43.0 & 72.5 & 17.8 \\
    \addlinespace[2pt]
    IC DiT-S & -- & 6.5 {\scriptsize[6.1, 6.9]} & 10.5 & 26.6 & 53.7 & 21.1 \\
    & DPS & 8.9 {\scriptsize[8.4, 9.4]} & 6.0 & 33.9 & 53.6 & 18.5 \\
    IC DiT-B & -- & 10.1 {\scriptsize[9.6, 10.6]} & 8.4 & 29.9 & 48.4 & 17.6 \\
    & DPS & 13.0 {\scriptsize[12.5, 13.6]} & 4.7 & 36.7 & 48.4 & 15.3 \\
    IC DiT-L & -- & 13.4 {\scriptsize[12.9, 14.0]} & 7.1 & 32.0 & 47.9 & 15.5 \\
    & DPS & 15.8 {\scriptsize[15.2, 16.4]} & 4.0 & 38.3 & 48.0 & 15.2 \\
    \midrule
    \multicolumn{7}{l}{\textit{BrickNet}} \\
    PT DiT-L & -- & 25.5 {\scriptsize[22.9, 28.0]} & 4.1 & 18.1 & 58.9 & 8.4 \\
    & DPS & 31.7 {\scriptsize[28.9, 34.4]} & 2.1 & 22.3 & 58.8 & 7.5 \\
    \addlinespace[2pt]
    IC DiT-S & -- & 19.4 {\scriptsize[16.8, 22.1]} & 6.3 & 15.2 & 34.9 & 11.3 \\
    & DPS & 24.4 {\scriptsize[21.4, 27.2]} & 3.9 & 18.6 & 34.8 & 9.7 \\
    IC DiT-B & -- & 26.1 {\scriptsize[23.0, 29.1]} & 5.2 & 16.6 & 31.5 & 9.3 \\
    & DPS & 31.7 {\scriptsize[28.5, 34.8]} & 3.2 & 19.7 & 31.4 & 7.8 \\
    IC DiT-L & -- & 30.5 {\scriptsize[27.4, 33.6]} & 4.4 & 17.8 & 31.0 & 8.0 \\
    & DPS & 35.1 {\scriptsize[32.0, 38.2]} & 2.8 & 20.6 & 30.9 & 7.3 \\
    \bottomrule
  \end{tabular}
\end{table}

\subsection{Image-Conditioned Reconstruction}
\Cref{tab:main} summarizes reconstruction quality across both evaluation sets.
Validity scales with model size: on BrickNet, unguided IC DiT-S/B/L reach 19.4\%, 26.1\%, and 30.5\%, and guidance adds 4.6--5.6 points at every size.
PT models scale likewise on BrickNet, from 13.6\% to 25.5\% (\cref{tab:full}).
Image conditioning reduces pose error, lowering MM from 58.9 to 31.0\,LDU on BrickNet (PT DiT-L to IC DiT-L).
ProcBrick follows the same trends at lower validity, as its assemblies are on average twice as large.
Finetuning on 50\% or 25\% of the renders lowers the BrickNet validity of IC DiT-S only moderately, from 19.4\% to 18.1\% and 16.2\% (\cref{sec:supp_ablations}).
Training IC DiT-S without the pretrained model, from random initialization and without the hand-off, lowers MM but increases the number of components and lowers BrickNet validity from 19.4\% to 2.3\% (\cref{sec:supp_ablations}).
Pretraining provides structural priors that improve connectivity even when individual part poses are slightly less accurate.

\begin{figure}[t]
  \centering
  \ifdefined\LaTeXML
    \includegraphics[width=\linewidth]{fig/plots/connectivity.svg}
  \else
    \includegraphics[width=\linewidth]{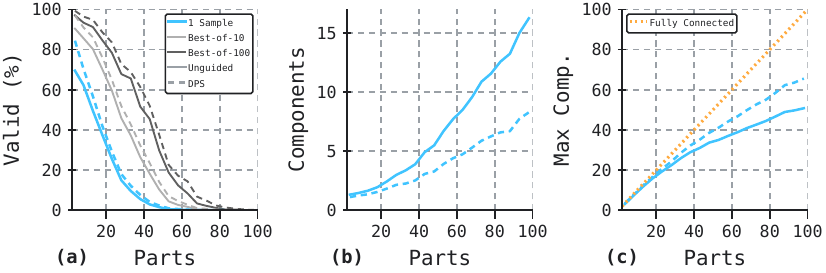}
  \fi
  \caption{\textbf{Validity and connectivity across part counts.}
    IC DiT-L on ProcBrick, unguided (solid) and guided (dashed).
    (a) Validity of one sample and of the best of 10 and 100.
    (b) Connected components; guidance removes 16\% of them for the smallest assemblies and about half beyond 80 parts.
    (c) Parts in the largest component, which grows with guidance from 51 to 66 for assemblies of 96--100 parts; the dotted line is a fully connected assembly.}
  \label{fig:connectivity}
\end{figure}

\subsection{Connectivity and Guidance}
Without guidance, the number of components grows with part count (\cref{fig:connectivity}).
Guidance reduces it at every part count, roughly halving it for the largest assemblies.
Guidance changes MM by at most 0.3\,LDU (\cref{tab:full}).
Guided samples also have fewer colliding pairs than unguided ones for every model (\cref{tab:main}).
Validity nonetheless falls with part count, with or without guidance, to below 1\% beyond about 50 parts for single samples and beyond 85 parts for the best of 100.
Large samples keep most parts in one component, but a single detached part or collision makes a sample invalid.

\subsection{Qualitative Results}
\begin{figure*}[t]
  \centering
  {\setlength{\parskip}{0pt}\offinterlineskip
    \ifdefined\rlcolw\else\newlength{\rlcolw}\newsavebox{\rlimg}\newsavebox{\rlstrut}\fi
    \sbox{\rlstrut}{\small\strut}\setlength{\rlcolw}{\dimexpr\ht\rlstrut+\dp\rlstrut+4pt\relax}
    \ifdefined\LaTeXML\def\rlrow#1#2{{\small #1}\par\includegraphics[width=\linewidth]{#2}\par}\else
    \def\rlrow#1#2{\sbox{\rlimg}{\includegraphics[width=\dimexpr\linewidth-\rlcolw\relax]{#2}}%
      \hbox to\linewidth{\rotatebox{90}{\hbox to\ht\rlimg{\hss\small\strut #1\hss}}\hfill\usebox{\rlimg}}\par}\fi
    \rlrow{Input}{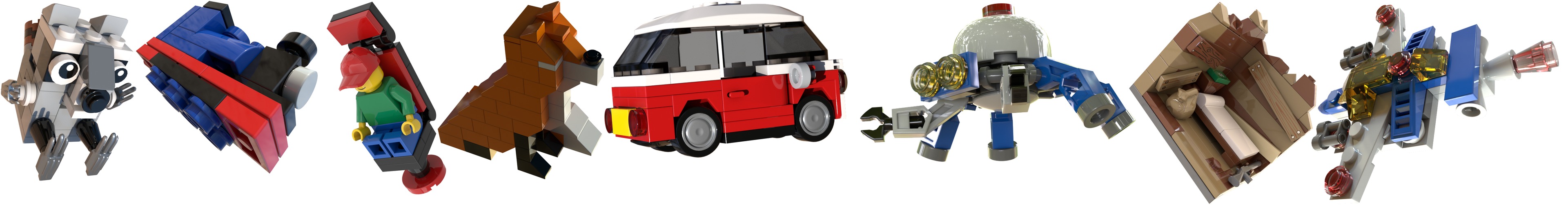}
    \rlrow{Output}{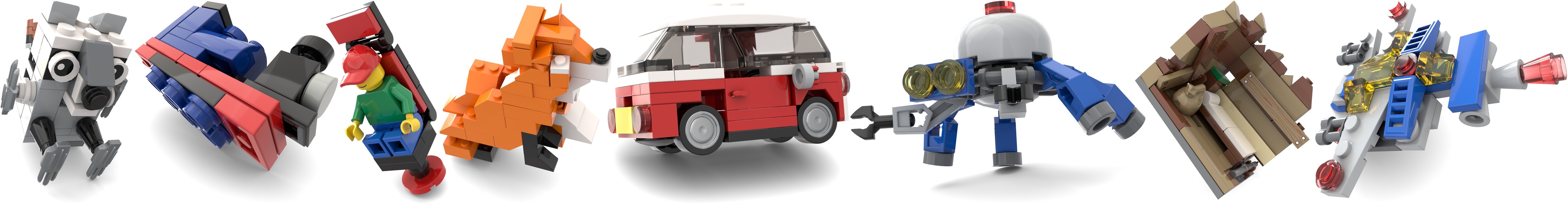}
    \rlrow{Input}{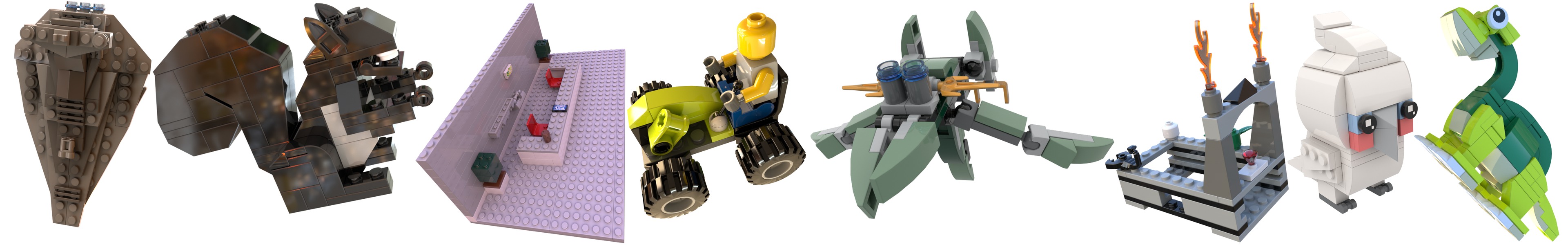}
    \rlrow{Output}{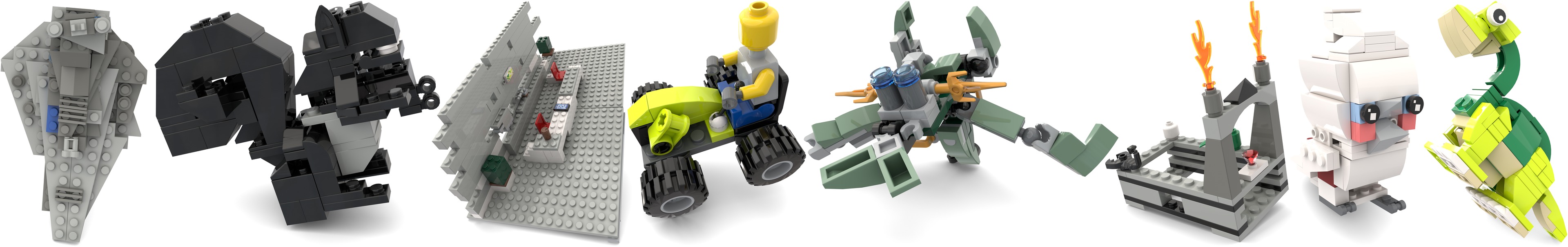}}
  \caption{\textbf{BrickNet reconstructions.}
    Outputs are selected from the guided IC DiT-L.
    As predictions are defined up to a global rigid transform, outputs are aligned to the ground truth as postprocessing.}
  \label{fig:bricknet-recon}
\end{figure*}

\begin{figure*}[t]
  \centering
  \ifdefined\LaTeXML
    {\small Input}\par
    \includegraphics[width=\linewidth]{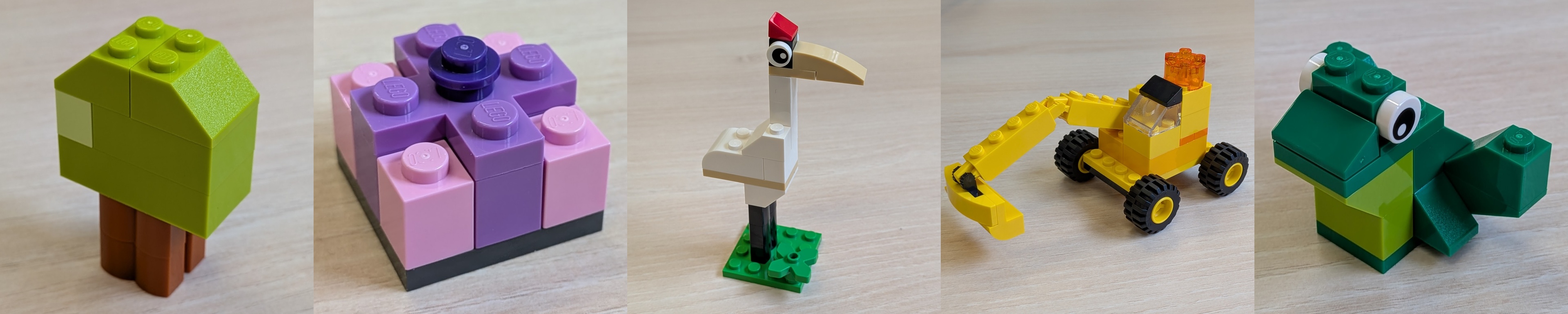}\par
    {\small Output}\par
    \includegraphics[width=\linewidth]{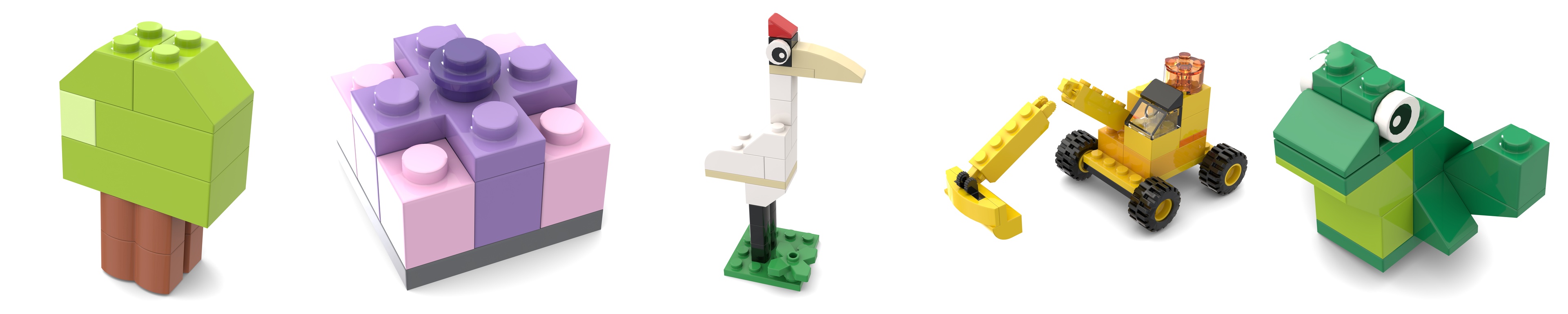}
  \else
    {\setlength{\parskip}{0pt}\offinterlineskip
      \ifdefined\rlcolw\else\newlength{\rlcolw}\newsavebox{\rlimg}\newsavebox{\rlstrut}\fi
      \ifdefined\rloverlap\else\newlength{\rloverlap}\newsavebox{\rlrowa}\newsavebox{\rlrowb}\fi
      \ifdefined\rlintop\else\newlength{\rlintop}\newlength{\rlinbot}\newlength{\rlouttop}\newlength{\rloutbot}\newlength{\rlw}\newlength{\rltrimt}\newlength{\rltrimb}\fi
      \setlength{\rloverlap}{0cm}
      \setlength{\rlintop}{0cm}\setlength{\rlinbot}{0cm}%
      \setlength{\rlouttop}{0cm}\setlength{\rloutbot}{0cm}%
      \sbox{\rlstrut}{\small\strut}\setlength{\rlcolw}{\dimexpr\ht\rlstrut+\dp\rlstrut+4pt\relax}
      \setlength{\rlw}{\dimexpr\linewidth-\rlcolw\relax}%
      \def\rlrow#1#2#3#4{\sbox{\rlimg}{\includegraphics{#2}}%
        \setlength{\rltrimt}{\dimexpr#3*\wd\rlimg/\rlw\relax}\setlength{\rltrimb}{\dimexpr#4*\wd\rlimg/\rlw\relax}%
        \edef\rlopts{width=\the\rlw,trim=0 \the\rltrimb\space 0 \the\rltrimt,clip}%
        \sbox{\rlimg}{\expandafter\includegraphics\expandafter[\rlopts]{#2}}%
        \hbox to\linewidth{\rotatebox{90}{\hbox to\ht\rlimg{\hss\small\strut #1\hss}}\hfill\usebox{\rlimg}}}
      \sbox{\rlrowa}{\rlrow{Input}{fig/images/rw_inputs.jpg}{\rlintop}{\rlinbot}}%
      \sbox{\rlrowb}{\rlrow{Output}{fig/images/rw_outputs.jpg}{\rlouttop}{\rloutbot}}%
      \vbox{\offinterlineskip
        \kern\dimexpr\ht\rlrowa+\dp\rlrowa-\rloverlap\relax
        \copy\rlrowb
        \kern-\dimexpr\ht\rlrowa+\dp\rlrowa-\rloverlap+\ht\rlrowb+\dp\rlrowb\relax
        \copy\rlrowa
        \kern\dimexpr\ht\rlrowb+\dp\rlrowb-\rloverlap\relax}\par}
  \fi
  \caption{\textbf{Real-world reconstructions.}
    Inputs are photographs of physical builds.
    Outputs are from guided IC DiT-L, and are rendered following a postprocessing global alignment to set the camera.}
  \label{fig:real-world}
\end{figure*}

Reconstruction examples on BrickNet are shown in \cref{fig:bricknet-recon}.
The model recovers visible structure and infers plausible placements for occluded parts.
\Cref{fig:real-world} shows reconstructions from photographs of physical builds, given their part inventories.

\section{Discussion and Limitations}
\label{sec:limitations}
While the connectivity-field guidance improves connectivity substantially, it does not guarantee fully connected or physically valid assemblies.
The field has no explicit collision term, so it reduces collisions only indirectly, by pulling parts into seats.
Achieving full validity may require better test-time optimization, for example with explicit collision terms or a discrete search over possibilities.

While our synthetic data appears to cover the test distribution fairly well, the model struggles in dense builds where parts look alike (\cref{fig:bricknet-recon}).
We suspect performance is limited by the distribution gap.

We assume that the inventory contains exactly the parts of the depicted assembly, and the method is not trained to handle missing, extra, or wrong parts.
Future work might consider how to handle this, but should consider more strongly the much more interesting problem of jointly inferring the object's partset along with the poses of the parts.
To make the method applicable to images captured in-the-wild, where the partset is not already known, such as to digitize a LEGO collection, interested readers might look to how co-generation is done in protein-generation literature such as in \citet{campbell2024generative} with mixed continuous--discrete denoising.

Finally, while we train on assemblies of up to 100 parts, real-world LEGO sets can contain thousands.
Scaling to larger assemblies will require addressing both computational cost and the increased ambiguity of inferring large hidden substructures from a single view.

\section{Conclusion}
\label{sec:conclusion}
Our work represents the first attempt to reconstruct a LEGO object given an image and its set of parts.

To achieve this, we first trained a set-conditioned $SE(3)$ flow model on procedurally generated assemblies (ProcBrick) to learn an adaptable assembly prior.
Second, we adapted this prior to single-image reconstruction using a fraction of the data, and despite training only on renders, demonstrated that the resulting model can produce plausible reconstructions of physical builds from photographs.
Finally, we introduced connectivity-field guidance, a test-time energy that pulls predictions toward connected assemblies during sampling, applicable without retraining, raising validity for every model at negligible pose cost and halving the number of disconnected components in large assemblies.

\begin{ack}
  This work was funded in part by the French government under management of Agence Nationale de la Recherche as part of the ``France 2030'' program, reference ANR-23-IACL-0008 (PR[AI]RIE-PSAI project).
  Cordelia Schmid would like to acknowledge the support by the K\"orber European Science Prize.
  We are grateful to Michael Black for his encouragement and support.
\end{ack}

\clearpage

{
  \small
  \bibliographystyle{plainnat}
  \bibliography{main}
}

\clearpage

\appendix

\section{Training Settings}\label{sec:supp_training}
\begin{table}[H]
  \caption{\textbf{Training settings.}
    EMA widths maximize the flow's log-likelihood on held-out assemblies.}
  \label{tab:training}
  \centering
  \footnotesize
  \setlength{\tabcolsep}{4pt}
  \begin{tabular}{@{}l l l@{}}
    \toprule
    & Pretraining (PT) & Finetuning (IC) \\
    \midrule
    Data & 100M 100-part assemblies & 1M rendered assemblies \\
    Part counts & prefixes of 2--100 & 2--100, balanced \\
    Steps, batch & 1M, 4{,}096 & 200k, 2{,}048 \\
    Muon LR (S/B/L) & $2.4/1.2/0.9 \times 10^{-3}$ & $7.1/3.5/2.7 \times 10^{-4}$ \\
    AdamW LR & $4 \times 10^{-4}$ & $1.4 \times 10^{-4}$ \\
    Flow time $t$ & logit-normal, $\mu{=}1.2$, $\sigma{=}1.5$~\citep{abramson2024alphafold3} & logit-normal, $\mu{=}0$, $\sigma{=}1.5$ \\
    Image encoder & -- & DINOv3-L layer 18, 512$\times$512 \\
    Resampler & -- & Perceiver, 256 latents, 6 layers \\
    Dropout & -- & image 10\%, materials 25\% \\
    Keypoint loss & -- & $\beta$-NLL, weight 10 \\
    \bottomrule
  \end{tabular}

  \vspace{6pt}
  \begin{tabular}{@{}l l@{}}
    \toprule
    Both stages & \\
    \midrule
    Optimizer & Muon~\citep{jordan2024muon} for hidden matrices, AdamW otherwise \\
    Schedule & inverse square root, 1k warmup \\
    Weights & post-hoc EMA~\citep{karras2024analyzing} \\
    DiT-S/B/L & width 384/768/1024, depth 12/12/24, 33.3M/131.7M/460.4M parameters in PT \\
    \bottomrule
  \end{tabular}
\end{table}

\section{Snapping}\label{sec:supp_snap}
After sampling, we snap each sample.
BrickNet detects its connections, and each detected connection is moved onto its exact seat if no part moves by more than 0.8\,mm/2\,LDU (rms) or 10$^\circ$, no connection is lost, and no collision is added.
It is otherwise left as-is.
All reported results are after snapping.
\begin{table}[H]
  \caption{\textbf{Effect of snapping.}
    Snapped minus raw, paired by sample, with 95\% CIs.
    Snapping raises validity and removes colliding pairs for every model while moving MM by at most 0.07\,LDU, and it adds less to guided samples.}
  \label{tab:snap}
  \centering
  \footnotesize
  \setlength{\tabcolsep}{3pt}
  \begin{tabular}{@{}l l r r r@{}}
    \toprule
    Model & Guid. & $\Delta$Valid & $\Delta$Coll. & $\Delta$MM \\
    \midrule
    \multicolumn{5}{l}{\textit{ProcBrick}} \\
    PT DiT-S & -- & 1.8 {\scriptsize[1.7, 1.9]} & $-$4.99 {\scriptsize[$-$5.07, $-$4.93]} & 0.02 {\scriptsize[0.02, 0.02]} \\
    & DPS & 0.8 {\scriptsize[0.8, 0.9]} & $-$5.10 {\scriptsize[$-$5.18, $-$5.02]} & 0.01 {\scriptsize[0.01, 0.01]} \\
    PT DiT-B & -- & 2.5 {\scriptsize[2.3, 2.7]} & $-$3.63 {\scriptsize[$-$3.69, $-$3.58]} & 0.06 {\scriptsize[0.06, 0.06]} \\
    & DPS & 1.2 {\scriptsize[1.1, 1.3]} & $-$4.47 {\scriptsize[$-$4.54, $-$4.40]} & 0.05 {\scriptsize[0.05, 0.05]} \\
    PT DiT-L & -- & 3.0 {\scriptsize[2.8, 3.2]} & $-$2.80 {\scriptsize[$-$2.85, $-$2.76]} & 0.07 {\scriptsize[0.07, 0.07]} \\
    & DPS & 1.9 {\scriptsize[1.8, 2.0]} & $-$5.66 {\scriptsize[$-$5.74, $-$5.57]} & 0.06 {\scriptsize[0.06, 0.06]} \\
    \addlinespace[2pt]
    IC DiT-S & -- & 2.1 {\scriptsize[2.0, 2.3]} & $-$4.57 {\scriptsize[$-$4.64, $-$4.50]} & 0.00 {\scriptsize[0.00, 0.00]} \\
    & DPS & 1.4 {\scriptsize[1.3, 1.5]} & $-$4.98 {\scriptsize[$-$5.06, $-$4.90]} & 0.00 {\scriptsize[0.00, 0.00]} \\
    IC DiT-B & -- & 3.0 {\scriptsize[2.8, 3.1]} & $-$3.48 {\scriptsize[$-$3.54, $-$3.43]} & 0.02 {\scriptsize[0.02, 0.02]} \\
    & DPS & 1.9 {\scriptsize[1.8, 2.0]} & $-$4.36 {\scriptsize[$-$4.43, $-$4.29]} & 0.02 {\scriptsize[0.02, 0.02]} \\
    IC DiT-L & -- & 3.2 {\scriptsize[3.1, 3.4]} & $-$2.56 {\scriptsize[$-$2.61, $-$2.51]} & 0.02 {\scriptsize[0.02, 0.02]} \\
    & DPS & 2.4 {\scriptsize[2.3, 2.6]} & $-$5.79 {\scriptsize[$-$5.89, $-$5.70]} & 0.02 {\scriptsize[0.02, 0.02]} \\
    \midrule
    \multicolumn{5}{l}{\textit{BrickNet}} \\
    PT DiT-S & -- & 4.9 {\scriptsize[4.1, 5.6]} & $-$2.84 {\scriptsize[$-$3.13, $-$2.58]} & 0.01 {\scriptsize[0.01, 0.01]} \\
    & DPS & 1.6 {\scriptsize[1.3, 2.0]} & $-$2.35 {\scriptsize[$-$2.60, $-$2.12]} & 0.01 {\scriptsize[0.01, 0.01]} \\
    PT DiT-B & -- & 6.9 {\scriptsize[6.0, 7.8]} & $-$2.27 {\scriptsize[$-$2.51, $-$2.04]} & 0.04 {\scriptsize[0.03, 0.04]} \\
    & DPS & 2.5 {\scriptsize[2.1, 2.9]} & $-$2.13 {\scriptsize[$-$2.37, $-$1.90]} & 0.03 {\scriptsize[0.03, 0.04]} \\
    PT DiT-L & -- & 6.4 {\scriptsize[5.5, 7.2]} & $-$1.75 {\scriptsize[$-$1.96, $-$1.55]} & 0.05 {\scriptsize[0.04, 0.06]} \\
    & DPS & 3.1 {\scriptsize[2.7, 3.6]} & $-$2.59 {\scriptsize[$-$2.91, $-$2.28]} & 0.04 {\scriptsize[0.04, 0.05]} \\
    \addlinespace[2pt]
    IC DiT-S & -- & 7.5 {\scriptsize[6.4, 8.7]} & $-$2.73 {\scriptsize[$-$2.98, $-$2.47]} & $-$0.01 {\scriptsize[$-$0.01, 0.00]} \\
    & DPS & 4.3 {\scriptsize[3.5, 5.1]} & $-$2.36 {\scriptsize[$-$2.61, $-$2.13]} & 0.00 {\scriptsize[$-$0.01, 0.00]} \\
    IC DiT-B & -- & 9.3 {\scriptsize[7.9, 10.6]} & $-$2.21 {\scriptsize[$-$2.45, $-$1.99]} & 0.00 {\scriptsize[$-$0.01, 0.00]} \\
    & DPS & 4.2 {\scriptsize[3.5, 4.9]} & $-$2.12 {\scriptsize[$-$2.38, $-$1.88]} & 0.00 {\scriptsize[0.00, 0.01]} \\
    IC DiT-L & -- & 7.0 {\scriptsize[5.9, 8.0]} & $-$1.70 {\scriptsize[$-$1.91, $-$1.51]} & $-$0.01 {\scriptsize[$-$0.01, 0.00]} \\
    & DPS & 4.4 {\scriptsize[3.7, 5.0]} & $-$2.65 {\scriptsize[$-$2.98, $-$2.35]} & 0.00 {\scriptsize[$-$0.01, 0.00]} \\
    \bottomrule
  \end{tabular}
\end{table}

\section{Connectivity Field Details}\label{sec:supp_connectivity}
The seat distance of a candidate seat $h$ between parts $a$ and $b$ is
\begin{equation}
  c_h = \min_{\theta,\, Q} \frac{\mu_a\, d_a^2 + \mu_b\, d_b^2}{\mu_a + \mu_b},
\end{equation}
where $d_a$ and $d_b$ are the displacements (\cref{eq:metric}) that bring the two parts onto the allowed relative transform with free coordinates $\theta$, and $Q$ is a common rigid motion of the pair, solved in closed form~\citep{horn1987closed}.
The taper $w_h$ is the quintic smoothstep~\citep{perlin2002improving} of the distance between the connector origins (minus, for axles, the largest admitted slide offset), equal to 1 below 12\,LDU and 0 beyond 16\,LDU.
The seat penalty is $q(c) = \tau H((c - \tau)/\tau)$ and the hinge is $\phi(x) = \Omega_0 H(x/\Omega_0)$ with $\Omega_0 = (N-1)\rho$, where $H(x) = x^3/(x^2+1)$ for $x > 0$ and $H(x) = 0$ otherwise.
The weak conductance between every pair of parts is $1/(16N)$\,LDU$^{-2}$.
We use $\rho = \tau = 10^{-8}$\,LDU$^2$.
On 1,000 held-out procedural assemblies, $E$ and its gradient are exactly zero.

\section{Guidance Hyperparameters}\label{sec:supp_guidance_sweep}
We select settings on ProcBrick evaluation assemblies stratified by part count, with 8 samples each, varying one factor at a time with shared noise, and check them on BrickNet.
Integrating rotations with the $\Delta t/(1-t)$ step of the training interpolant instead of the exponential schedule more than halves validity on this subset (PT DiT-L: 7.0\% vs.\ 17.2\% unguided).
\begin{table}[H]
  \caption{\textbf{Guidance hyperparameters.}
    Validity (\%) of guided samples before snapping.
    Each change moves validity by at most 2.1 points.
    Top: changes from the initial setting.
    Bottom: changes from the final setting.}
  \label{tab:dps_choices}
  \centering
  \footnotesize
  \setlength{\tabcolsep}{4pt}
  \begin{tabular}{@{}l r r@{}}
    \toprule
    Setting & PT DiT-L & PT DiT-S \\
    \midrule
    initial: $\gamma{=}2$, clip 1\,LDU, $t_{\mathrm{start}}{=}0.7$, $p{=}4$, taper 8--12\,LDU & 28.2 & 21.0 \\
    \quad $\gamma = 1$ / $4$ & 26.7 / 26.8 & 19.7 / 20.6 \\
    \quad clip 0.5 / 2\,LDU & 28.5 / 27.8 & 21.2 / 21.4 \\
    \quad $t_{\mathrm{start}} = 0.6$ / $0.8$ & 27.5 / 28.2 & 20.9 / 21.5 \\
    \midrule
    \textbf{final}: taper 12--16\,LDU & \textbf{28.6} & \textbf{21.5} \\
    \quad no clip & 26.5 & 21.5 \\
    \quad $t_{\mathrm{start}} = 0$ & 26.8 & -- \\
    \bottomrule
  \end{tabular}
\end{table}

\section{Full Results}\label{sec:supp_full}
\begin{table}[H]
  \caption{\textbf{Full results.}
    Validity at collision allowances of 1.1 and 1.0\,LDU, and MM, with 100 samples per structure.
    Validity at 1.0\,LDU is nearly identical, and guidance changes MM by at most 0.3\,LDU.}
  \label{tab:full}
  \centering
  \footnotesize
  \setlength{\tabcolsep}{4pt}
  \begin{tabular}{@{}l l r r r@{}}
    \toprule
    Model & Guid. & V@1.1 & V@1.0 & MM \\
    \midrule
    \multicolumn{5}{l}{\textit{ProcBrick}} \\
    PT DiT-S & -- & 5.5 & 5.5 & 72.45 \\
    & DPS & 7.7 & 7.7 & 72.13 \\
    PT DiT-B & -- & 7.8 & 7.7 & 72.38 \\
    & DPS & 10.9 & 10.8 & 72.21 \\
    PT DiT-L & -- & 10.3 & 10.3 & 72.56 \\
    & DPS & 13.5 & 13.3 & 72.53 \\
    \addlinespace[2pt]
    IC DiT-S & -- & 6.5 & 6.5 & 53.73 \\
    & DPS & 8.9 & 8.8 & 53.64 \\
    IC DiT-B & -- & 10.1 & 10.1 & 48.38 \\
    & DPS & 13.0 & 12.9 & 48.37 \\
    IC DiT-L & -- & 13.4 & 13.4 & 47.93 \\
    & DPS & 15.8 & 15.6 & 47.96 \\
    \midrule
    \multicolumn{5}{l}{\textit{BrickNet}} \\
    PT DiT-S & -- & 13.6 & 13.4 & 58.41 \\
    & DPS & 18.0 & 17.9 & 58.12 \\
    PT DiT-B & -- & 19.8 & 19.7 & 58.65 \\
    & DPS & 25.9 & 25.7 & 58.49 \\
    PT DiT-L & -- & 25.5 & 25.3 & 58.85 \\
    & DPS & 31.7 & 31.5 & 58.78 \\
    \addlinespace[2pt]
    IC DiT-S & -- & 19.4 & 19.2 & 34.90 \\
    & DPS & 24.4 & 24.2 & 34.82 \\
    IC DiT-B & -- & 26.1 & 26.0 & 31.47 \\
    & DPS & 31.7 & 31.5 & 31.43 \\
    IC DiT-L & -- & 30.5 & 30.3 & 30.95 \\
    & DPS & 35.1 & 34.8 & 30.94 \\
    \bottomrule
  \end{tabular}
\end{table}

\section{Inference Time}\label{sec:supp_timing}
Guidance makes one sample 1.5--3.4 times slower, almost entirely from work on the GPU rather than the energy; with batching on ProcBrick, it slows the IC models by at most 1.15 times.
From 2--10 to 71--100 parts, time per sample rises by at most 1.4 times.

\begin{table}[H]
  \caption{\textbf{Inference time.}
    Seconds to sample one assembly on one H100 GPU.}
  \label{tab:timing}
  \centering
  \footnotesize
  \setlength{\tabcolsep}{4pt}
  \begin{tabular}{@{}l r r r r@{}}
    \toprule
    & \multicolumn{2}{c}{Seconds per sample} & \multicolumn{2}{c}{Samples per second} \\
    \cmidrule(lr){2-3} \cmidrule(lr){4-5}
    Model & Unguided & DPS & Unguided & DPS \\
    \midrule
    PT DiT-S & 0.09 & 0.30 & 919 & 553 \\
    PT DiT-B & 0.11 & 0.33 & 374 & 286 \\
    PT DiT-L & 0.25 & 0.62 & 125 & 98 \\
    \addlinespace[2pt]
    IC DiT-S & 0.39 & 0.68 & 106 & 97 \\
    IC DiT-B & 0.51 & 0.75 & 53 & 47 \\
    IC DiT-L & 1.00 & 1.49 & 19 & 17 \\
    \bottomrule
  \end{tabular}
\end{table}

\section{Ablations}\label{sec:supp_ablations}
\begin{table}[H]
  \caption{\textbf{Ablations.}
    Variants of IC DiT-S, without guidance.
    Training without the pretrained model collapses validity, and less rendered data lowers it moderately.
    10 samples per structure (100 for the reference rows).}
  \label{tab:ablations}
  \centering
  \footnotesize
  \setlength{\tabcolsep}{3pt}
  \begin{tabular}{@{}l r r r@{}}
    \toprule
    Variant & Valid & Comp. & MM \\
    \midrule
    \multicolumn{4}{l}{\textit{ProcBrick, IC DiT-S}} \\
    IC DiT-S & 6.5 {\scriptsize[6.1, 6.9]} & 10.46 & 53.73 \\
    from scratch & 1.2 {\scriptsize[1.0, 1.4]} & 32.15 & 50.43 \\
    50\% renders & 6.3 {\scriptsize[5.9, 6.7]} & 10.62 & 54.13 \\
    25\% renders & 5.8 {\scriptsize[5.4, 6.3]} & 11.07 & 55.24 \\
    \midrule
    \multicolumn{4}{l}{\textit{BrickNet, IC DiT-S}} \\
    IC DiT-S & 19.4 {\scriptsize[16.8, 22.1]} & 6.25 & 34.90 \\
    from scratch & 2.3 {\scriptsize[1.3, 3.5]} & 16.51 & 33.83 \\
    50\% renders & 18.1 {\scriptsize[15.6, 20.7]} & 6.34 & 35.66 \\
    25\% renders & 16.2 {\scriptsize[13.7, 18.8]} & 6.58 & 37.76 \\
    \bottomrule
  \end{tabular}
\end{table}

\section{ProcBrick Details}\label{sec:supp_data}

\noindent\textbf{Generation.}
A placement is rejected if two parts interpenetrate by more than 1\,LDU, unless they are joined by a fixed connection.
Part-type frequencies are tilted toward larger parts and calibrated per connector family.
The seed part follows the occurrence frequencies of part types in human-designed models.
The acceptance test penalizes drops in packing density, and its targets for connector joins and connected neighbors per added part span roughly the median to the upper quantiles of human-designed models.

\noindent\textbf{Splits.}
All 100M pretraining assemblies pass the BrickNet collision check on every prefix.
From a separate pool, we draw disjoint sets of 1M image-training, 9,900 evaluation, and 1,000 geometry-validation assemblies.

\noindent\textbf{Appearance and rendering.}
Each part's material is one of 260 renderer materials, drawn by a policy fitted to the palettes of 53,431 human-designed models.
We render at 1024$\times$1024 on a transparent background, lit by one of 964 environment maps or, for 10\% of images, a studio setup.
During finetuning, we composite renders onto ImageNet~\citep{deng2009imagenet} or plain backgrounds and apply random camera roll, zoom, and offset, random erasing, photometric jitter, and sensor noise.

\end{document}